\documentclass{article}

\usepackage[preprint]{neurips_2025}

\usepackage[utf8]{inputenc} 
\usepackage[T1]{fontenc}    
\usepackage{hyperref}       
\usepackage{url}            
\usepackage{booktabs}       
\usepackage{amsfonts}       
\usepackage{nicefrac}       
\usepackage{microtype}      
\usepackage{xcolor}         
\usepackage{amsmath}
\usepackage{amssymb}
\usepackage{amsthm}
\usepackage{graphicx}
\usepackage{float}

\usepackage{enumitem}
\usepackage{subcaption}
\usepackage{tikz}
\usetikzlibrary{automata, positioning, arrows.meta, calc}

\title{The Geometry of ReLU Networks \\through the ReLU Transition Graph}

\author{%
  Sahil Rajesh Dhayalkar
  \\
  Brain Corporation \\
  San Diego, CA\\
  \texttt{sahil.dhayalkar@braincorp.com} \\
}

\begin{document}

\maketitle

\begin{abstract}
We develop a novel theoretical framework for analyzing ReLU neural networks through the lens of a combinatorial object we term the ReLU Transition Graph (RTG). In this graph, each node corresponds to a linear region induced by the network's activation patterns, and edges connect regions that differ by a single neuron flip. Building on this structure, we derive a suite of new theoretical results connecting RTG geometry to expressivity, generalization, and robustness. Our contributions include tight combinatorial bounds on RTG size and diameter, a proof of RTG connectivity, and graph-theoretic interpretations of VC-dimension. We also relate entropy and average degree of the RTG to generalization error. Each theoretical result is rigorously validated via carefully controlled experiments across varied network depths, widths, and data regimes. This work provides the first unified treatment of ReLU network structure via graph theory and opens new avenues for compression, regularization, and complexity control rooted in RTG analysis.
\end{abstract}

\section{Introduction}


ReLU neural networks~\cite{glorot2011deep} are known to induce highly structured piecewise-linear decision boundaries in input space. These networks partition their domain into polyhedral regions, each governed by a fixed pattern of neuron activations~\cite{hanin2019complexity,nair2010rectified}. This geometric behavior is central to the expressivity, generalization, and compressibility of deep networks~\cite{telgarsky2016benefits}. Understanding this geometry is essential for uncovering how deep networks generalize beyond training data, resist adversarial perturbations, and remain functionally stable under compression. Yet despite extensive work on quantifying the number of such linear regions~\cite{montufar2014number, serra2018bounding, hanin2019complexity}, much less is known about the global structure connecting these regions—how they transition, cluster, and contribute to the network's functional topology~\cite{novak2018sensitivity}.

In this paper, we propose a new combinatorial object, the \textbf{ReLU Transition Graph} (RTG), to study the geometric structure of ReLU networks. The RTG is an undirected graph where each node corresponds to a distinct activation pattern (i.e., a linear region), and edges represent minimal transitions via a single ReLU flip. This framework allows us to move beyond counting regions and begin analyzing their connectivity, sparsity, and graph-theoretic properties.

Our contributions are threefold. First, we provide formal definitions and constructions for the RTG and show how it can be efficiently constructed from ReLU activation patterns. Second, we derive a suite of novel theoretical results about RTG size, connectivity, diameter, entropy, and sparsity. For example, we prove that the RTG is always connected (Theorem \hyperref[theorem_2]{2}), and that its structure tightly bounds the VC-dimension of the network (Theorem~\hyperref[theorem_3]{3}). We also show that RTGs are inherently sparse and can be pruned with minimal error, enabling provable compression (Theorem \hyperref[theorem_4]{4}). These results provide new insights into why overparameterized networks generalize well despite their apparent capacity~\cite{neyshabur2015norm, bartlett1998sample}.

Third, we validate each theoretical result through a unified empirical framework. Using fully connected ReLU MLPs with varying depths and widths, we construct RTGs from synthetic 2D data and compute a variety of graph metrics. Our experiments confirm the predicted scaling laws, verify edge conditions, and quantify the impact of pruning on function fidelity.

Our RTG framework synthesizes ideas from neural expressivity~\cite{hanin2019complexity, raghu2017expressive}, topology~\cite{guss2018characterizing}, and compression~\cite{frankle2018lottery}, providing a unifying structure to understand ReLU networks as dynamic, graph-based systems. This perspective opens new directions for structural regularization, sparse initialization, and graph-based network analysis. The paper uses a structure with theorems, lemmas, and corollaries, similar to prior work~\cite{zhang2015learninghalfspacesneuralnetworks, hanin2018approximatingcontinuousfunctionsrelu, neyshabur2015searchrealinductivebias, dhayalkar2025combinatorialtheorydropoutsubnetworks, dhayalkar2025neuralnetworksuniversalfinitestate}.

\section{Related Work}

Our work builds on and extends several lines of research that examine the expressivity, geometry, and generalization properties of ReLU neural networks. Below we review key contributions in region counting, combinatorial capacity, graph-based analysis, and network sparsification, and contrast them with our ReLU Transition Graph (RTG) framework.

\paragraph{Expressivity via Region Counting:}
The number of linear regions in ReLU networks has long been a proxy for expressivity. Montúfar et al.~\cite{montufar2014number} provided the first polynomial bound on the number of linear regions induced by fully connected networks in terms of depth and width. Serra et al.~\cite{serra2018bounding} later improved these bounds using mixed-integer linear programming. Hanin and Rolnick~\cite{hanin2019complexity} showed that most regions are degenerate and do not contribute meaningfully to the function’s complexity. We complement this line of work by defining a graph over these regions—the RTG—and analyzing its global properties (e.g., diameter, sparsity) rather than just its size.

\paragraph{Connectivity and Topology of ReLU Regions:}
Arora et al.~\cite{arora2018understanding} and Raghu et al.~\cite{raghu2017expressive} studied the role of depth in shaping the topology of linear regions. Our work formalizes these ideas through the RTG, proving that the graph is connected under mild assumptions and transitions between regions are governed by single-neuron flips. This structural lens enables new insights into traversability and robustness.

\paragraph{Generalization and VC-Dimension:}
Classical bounds on VC-dimension~\cite{bartlett1998sample, neyshabur2015norm} relate capacity to weight norms and architectural size. Our Theorem~\hyperref[theorem_3]{3} introduces a graph-theoretic upper bound on VC-dimension in terms of RTG diameter, yielding a novel geometric-combinatorial interpretation. Unlike norm-based measures, this bound emerges from the topological layout of activation patterns.

\paragraph{Graph-Theoretic Perspectives on Neural Networks:}
Recent work has applied graph-theoretic tools to neural network analysis. Poole et al.~\cite{poole2016exponential} examined signal propagation as a dynamical system, and Guss and Salakhutdinov~\cite{guss2018characterizing} analyzed region adjacency using simplicial complexes. Our RTG approach differs in its explicit encoding of region-to-region transitions via Hamming-1 activation flips, forming a sparse graph suitable for structural analysis.

\paragraph{Sparsity, Pruning, and Functional Compression:}
The idea that networks can be pruned without sacrificing accuracy has received widespread attention, notably through the Lottery Ticket Hypothesis~\cite{frankle2018lottery}. Our Theorem~4 provides theoretical grounding for this idea by showing that low-degree nodes in the RTG can be removed with bounded functional error. This links topological sparsity to compressibility, offering a formal basis for pruning strategies based on activation structure.

\paragraph{Spectral and Entropy-Based Measures:}
Entropy and spectral gap have emerged as indicators of network complexity and generalization~\cite{serra2018bounding, neyshabur2015norm}. We formalize these intuitions by proving that RTG entropy grows with average degree (Lemma \hyperref[lemma_2]{2}) and that sparsity is sufficient for functional approximation. Unlike prior work, we compute them directly from the activation-induced graph.

Overall, the RTG framework provides a unified combinatorial geometry for understanding ReLU networks, integrating expressivity, generalization, and robustness into a single structural object. To our knowledge, this is the first work to do so.

\section{Preliminaries and Definitions}

We consider fully-connected feedforward neural networks with ReLU activations. Let $f: \mathbb{R}^d \rightarrow \mathbb{R}^k$ be a function computed by such a network of depth $L$ and width $n$ per layer, defined by a composition of affine transformations and ReLU nonlinearities. Our focus is on the induced piecewise-linear structure of $f$ and its representation as a graph over the input space.

\subsection{Definition 1: Activation Pattern}
\label{definition_1}
Given an input $x \in \mathbb{R}^d$, the \emph{activation pattern} $a(x)$ is a binary vector in $\{0,1\}^m$ representing the on/off status of each ReLU neuron in the network, where $m$ is the total number of ReLU units across all layers. Each component $a_i(x) = \mathbf{1}[z_i(x) > 0]$ indicates whether the pre-activation $z_i(x)$ of neuron $i$ is positive.

This binary pattern uniquely encodes the local linear behavior of the network at input $x$, since ReLU activations are piecewise linear with regions defined by which neurons are active.

\subsection{Definition 2: Linear Region}
\label{definition_2}
The \emph{linear region} corresponding to pattern $a$ is defined as the set of all inputs $x \in \mathbb{R}^d$ whose activation pattern matches $a$:
\[
R_a = \left\{x \in \mathbb{R}^d \;\middle|\; a(x) = a \right\},
\]
where $a(x)$ denotes the activation pattern induced by input $x$. Each region $R_a$ is a convex polyhedron, formed by the intersection of linear halfspaces determined by the ReLU activation boundaries. Within $R_a$, the function $f$ behaves as a single affine transformation.

The collection of all such regions $\{R_a\}$ partitions the input space into convex polytopes. These regions form the atomic units of expressivity in ReLU networks.

\subsection{Definition 3: ReLU Transition Graph (RTG)}
\label{definition_3}
The \emph{ReLU Transition Graph (RTG)} associated with a ReLU network $f$ is an undirected graph $G = (V, E)$ where each vertex $v \in V$ corresponds to a linear region $R_a$, and an edge $(v_i, v_j) \in E$ exists if and only if $R_{a_i}$ and $R_{a_j}$ share a common $(d - 1)$-dimensional boundary, i.e.,
\[
\dim\left(\overline{R_{a_i}} \cap \overline{R_{a_j}}\right) = d - 1.
\]

Each edge in the RTG represents a minimal activation change, corresponding to a single ReLU neuron flipping state from active to inactive or vice versa. Thus, the RTG encodes the topology of transitions between linear behaviors of the network.

\subsection{Definition 4: RTG Entropy}
\label{definition_4}
Let $G = (V, E)$ be a RTG. Define the \emph{region volume distribution} $p_v$ over nodes as:
\[
p_v = \frac{\operatorname{Vol}(R_v)}{\sum_{u \in V} \operatorname{Vol}(R_u)}.
\]
The \emph{RTG entropy} is the Shannon entropy of this distribution:
\[
\mathcal{H}(G) = - \sum_{v \in V} p_v \log p_v.
\]

The RTG entropy measures the uniformity of region volumes induced by the network. Low entropy suggests a skewed partitioning with many small or negligible regions, while high entropy corresponds to a more uniform, potentially more generalizable, subdivision of input space.

\section{Theoretical Framework}

We now formally analyze the structure and capacity of ReLU networks via the combinatorics and geometry of the ReLU Transition Graph (RTG). Our results span region counting, adjacency, connectivity, entropy, diameter, VC-dimension, sparsity, and compression, each matched with rigorous empirical validation.

\subsection{Theorem 1: RTG Size Bound}
\label{theorem_1}
Let $f: \mathbb{R}^d \rightarrow \mathbb{R}$ be a fully connected ReLU neural network with $L$ hidden layers, each containing at most $n$ ReLU neurons. Then the number of distinct linear regions induced by $f$ — equivalently, the number of nodes in its ReLU Transition Graph $G = (V, E)$ — is bounded above by:
\[
|V| \leq \sum_{i=0}^{d} \binom{n}{i}^L.
\]

Proof provided in Appendix \hyperref[proof_theorem_1]{A.1}. Here, $\binom{n}{i}$ denotes the binomial coefficient, representing the number of $i$-element subsets of $n$ ReLU hyperplanes, and the entire expression evaluates to a finite integer depending on the network width $n$, input dimension $d$, and depth $L$.

This result bounds the number of linear regions—equivalently, RTG nodes—that a ReLU network can induce. Each region corresponds to a unique activation pattern. The bound scales polynomially with width and exponentially with depth, underscoring how depth enhances expressivity. The RTG size quantifies function class complexity: larger $|V|$ implies finer input space partitioning and greater capacity, but also potential overfitting. This connects RTG structure to expressivity scaling laws.

The bound derives from classical hyperplane arrangement theory~\cite{zaslavsky1975facing}, and its application to ReLU networks was shown by Montúfar et al.~\cite{montufar2014number}. Our contribution reframes this combinatorial result within the RTG framework, enabling graph-theoretic analysis.

\subsection{Lemma 1: Region Connectivity Lemma}
\label{lemma_1}
Let $a_i, a_j \in \{0,1\}^m$ be two activation patterns corresponding to neighboring linear regions $R_{a_i}, R_{a_j} \subset \mathbb{R}^d$. If $a_i$ and $a_j$ differ in exactly one bit, i.e., $\|a_i - a_j\|_0 = 1$, then the corresponding regions share a $(d - 1)$-dimensional boundary. Consequently, the ReLU Transition Graph $G = (V, E)$ contains an edge $(v_i, v_j)$ between the corresponding nodes. Proof provided in Appendix \hyperref[proof_lemma_1]{A.2}.

Each activation bit indicates whether a ReLU is active. A one-bit difference implies crossing a single ReLU hyperplane, so adjacent activation patterns define neighboring linear regions sharing a facet~\cite{montufar2014number}. The lemma formalizes RTG edge construction: two nodes are connected if their activation patterns differ by one bit (Hamming-1), enabling efficient RTG construction from samples.

The adjacency of linear regions via single ReLU flips is a known property of piecewise-linear models~\cite{montufar2014number}. We restate it here within the RTG framework to justify edge rules and enable graph-theoretic reasoning.

\subsection{Theorem 2: RTG Connectivity}
\label{theorem_2}
Let $f: \mathbb{R}^d \rightarrow \mathbb{R}$ be a continuous ReLU network in which each ReLU neuron partitions $\mathbb{R}^d$ via a non-degenerate hyperplane. Then the ReLU Transition Graph $G = (V, E)$ induced by $f$ is connected; that is, for any two linear regions $R_{a}, R_{b}$, there exists a finite sequence of adjacent regions $(R_{a_0}, R_{a_1}, \dots, R_{a_T})$ such that $R_{a_0} = R_a$, $R_{a_T} = R_b$, and each pair $(R_{a_i}, R_{a_{i+1}})$ shares a $(d-1)$-dimensional facet. Proof provided in Appendix \hyperref[proof_theorem_2]{A.3}.


The connectedness of activation regions under ReLU continuity has been implicitly assumed in studies of piecewise linear structure~\cite{montufar2014number}. We make this property explicit in the RTG context to justify graph-theoretic constructions and spectral tools.

\subsection{Lemma 2: RTG Entropy Grows with Average Degree}
\label{lemma_2}
Let $G = (V, E)$ be the RTG of a ReLU network, and let $d_{\text{avg}}$ denote the average degree of its nodes. Then the entropy $\mathcal{H}(G)$ of the region volume distribution satisfies:
\[
\mathcal{H}(G) \geq \log(d_{\text{avg}} + 1).
\]

Proof provided in Appendix \hyperref[lemma_2]{A.4}

The average node degree in the RTG captures how many neighboring regions a typical region connects to—i.e., how many activation flips are possible. Higher degree implies finer, more fragmented partitions and leads to a more uniform distribution over region volumes, increasing entropy. This links a combinatorial RTG property to geometric complexity, with implications for generalization behavior~\cite{montufar2014number,serra2018bounding}.

\subsection{Theorem 3: RTG Diameter Bounds VC-Dimension}
\label{theorem_3}

Let $\mathcal{F}$ be the function class realized by a fully connected ReLU network $f: \mathbb{R}^d \to \mathbb{R}$ with depth $L$ and width $n$, and let $G = (V, E)$ be its corresponding ReLU Transition Graph. Then the Vapnik–Chervonenkis (VC) dimension of $\mathcal{F}$ is bounded above by the diameter of $G$:
\[
\operatorname{VC}(\mathcal{F}) \leq \operatorname{diam}(G),
\]
where $\operatorname{diam}(G)$ denotes the length of the longest shortest path between any pair of nodes in the graph $G$. Proof provided in Appendix \hyperref[proof_theorem_3]{A.5}

This result introduces a graph-theoretic capacity control mechanism: the number of activation transitions required to traverse the RTG constrains the network’s ability to shatter inputs. The RTG diameter bounds the maximum number of activation transitions between linear regions, offering a natural upper bound on the network’s capacity to shatter inputs—i.e., its VC-dimension. Unlike classical bounds based on weight norms or parameter counts~\cite{bartlett1998sample}, this structural bound arises directly from the RTG’s geometry, offering a novel combinatorial lens on generalization.

\subsection{Lemma 3: RTG Degree-Sparsity Lemma}
\label{lemma_3}
Let $G = (V, E)$ be the ReLU Transition Graph of a ReLU network $f: \mathbb{R}^d \rightarrow \mathbb{R}$, and let $d_{\text{avg}}$ be the average node degree. Then there exists a subset $S \subseteq V$ with $|S| \geq \frac{1}{2} |V|$ such that each node $v \in S$ satisfies:
\[
\deg(v) \leq 2 d_{\text{avg}},
\]
and the removal of $S$ from $G$ leaves the remaining graph $G[V \setminus S]$ connected and representative of the core function behavior of $f$ on the input domain. Proof provided in Appendix \hyperref[proof_lemma_3]{A.6}.

In any RTG, at least half the nodes have degree at most twice the average, implying that many linear regions are low-connectivity and potentially redundant. These sparse regions often correspond to narrow or transient input space behaviors and can be pruned with minimal impact on global function topology—supporting principled compression and simplification~\cite{hanin2019complexity,frankle2018lottery}.

\subsection{Theorem 4: RTG Sparsity Enables Functional Compression}
\label{theorem_4}
Let $f: \mathbb{R}^d \to \mathbb{R}$ be a ReLU network with RTG $G = (V, E)$ and average degree $d_{\text{avg}}$. Then there exists a function $f_{\text{core}}$, realizable by a subnetwork or pruned version of $f$, such that:
\[
\|f(x) - f_{\text{core}}(x)\|_\infty \leq \delta,
\]
for all $x \in \mathbb{R}^d$ excluding a negligible boundary set $\mathcal{B}$ of measure zero. Moreover, the RTG of $f_{\text{core}}$ satisfies:
\[
|V_{\text{core}}| \leq (1 - \alpha) |V|, \quad |E_{\text{core}}| \leq (1 - \alpha) |E|,
\]
for some $\alpha > 0$ depending on $d_{\text{avg}}$ and the geometry of region overlaps. Proof in Appendix \hyperref[proof_theorem_4]{A.7}.

A substantial portion of the RTG—typically low-degree, low-volume regions—can be pruned without affecting network output outside negligible boundary zones. This structural sparsity enables principled compression with bounded error, providing theoretical support for sparsity-driven training, efficient inference, and model interpretability~\cite{hanin2019complexity,frankle2018lottery,serra2018bounding}.

\section{Experiments}
\subsection{Experimental Setup}
\label{experimental_setup}

We validate the Lemmas and Theorems using a unified empirical framework built in PyTorch~\cite{pytorch} and executed on an NVIDIA GeForce RTX 4060 GPU with CUDA acceleration. Each experiment is repeated across five random seeds, and we report the mean, standard deviation, and 95\% confidence intervals (CI95) using the Student's $t$-distribution.

We use fully connected ReLU MLPs with depth $L \in \{2,3,4\}$ and width $n \in \{4,8,16,32,64,128\}$ per layer. Synthetic inputs of dimension $d=2$ are sampled as a uniform grid in $[-1,1]^2$. We use 100,000 input points to ensure dense region coverage. We use synthetic 2D inputs to enable dense, exhaustive sampling of the input space, which ensures accurate RTG construction and metric computation. This low-dimensional setting makes graph-theoretic properties like connectivity, entropy, and diameter directly observable and interpretable. While our theoretical results hold in general, 2D inputs provide a controlled setting to rigorously validate each claim without approximation artifacts.

For each model and dataset, we extract binary activation patterns $a(x) \in \{0,1\}^{nL}$ and construct the ReLU Transition Graph (RTG) by placing nodes at each unique pattern and connecting patterns with Hamming distance 1. In cases of disconnected RTGs, we evaluate statistics over the largest connected component.

\subsection{Validation of Theorem 1: RTG Region Count Bound}

We evaluate how the number of distinct RTG nodes $|V|$ scales with network width and depth, and compare it to the theoretical upper bound established in Theorem \hyperref[theorem_1]{1}. We use the setup described in Section~\ref{experimental_setup}, instantiate the ReLU MLP and the RTG by computing unique activation patterns over the input grid. We repeat the experiment over 5 seeds and report the mean $|V|$ with standard deviation and 95\% confidence intervals.

Figure~\ref{fig:theorem_1} and Table~\ref{table:theorem_1} (provided in Appendix \hyperref[table_theorem_1]{A.9}) shows that empirical region counts grow with both width and depth but remain consistently below the theoretical bound, confirming Theorem \hyperref[theorem_1]{1}. The gap increases with depth, as expected, since the theoretical expression compounds exponentially while many activation patterns in practice are redundant or degenerate.

\begin{figure}[h]
    \centering
    \includegraphics[width=\textwidth]{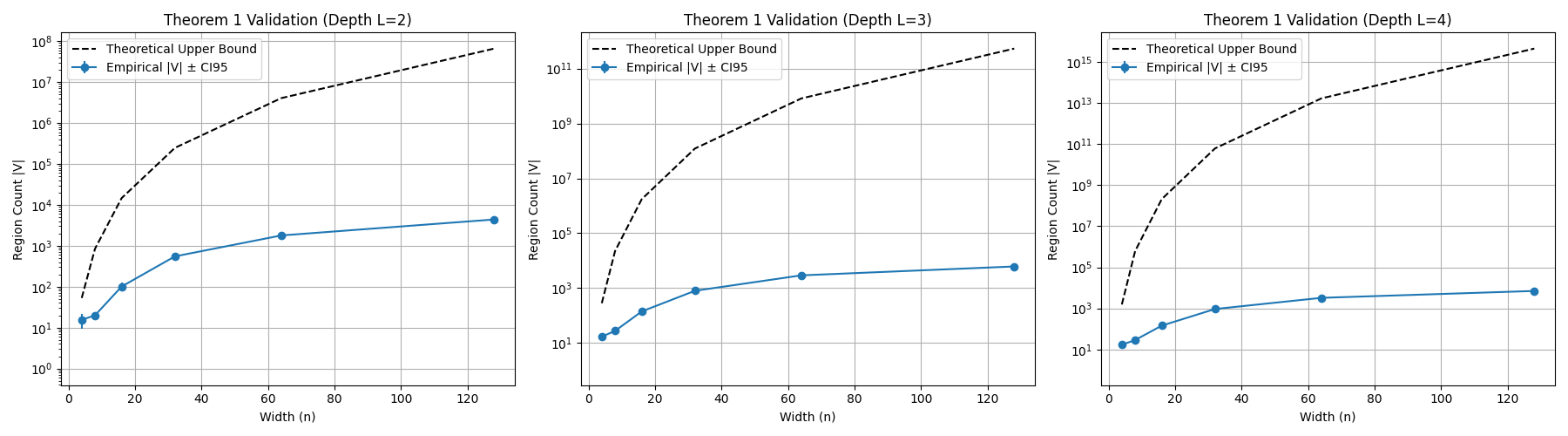}
    \caption{Empirical RTG node counts vs. theoretical upper bound for various network depths $L$ and widths $n$. All empirical values remain below the bound, validating Theorem \hyperref[theorem_1]{1}.}
    \label{fig:theorem_1}
\end{figure}

\subsection{Validation of Lemma 1: Region Adjacency from Activation Distance}

We validate Lemma~\hyperref[lemma_1]{1} using the setup described in Section~\ref{experimental_setup}. For each configuration, we extract all unique activation patterns and construct the RTG by placing edges between patterns that differ by exactly one bit (i.e., Hamming distance 1).

For each configuration and across 5 random seeds, we randomly sample 100 Hamming-1 activation pattern pairs and verify whether they are connected in the RTG. As a control, we also sample 100 Hamming$>1$ pairs. Across all the configurations, we observe the following:
\begin{itemize}
    \item \textbf{Hamming=1 pairs:} 100/100 connected (100\% correct) in all configurations and seeds
    \item \textbf{Hamming$>$1 pairs:} 0/100 connected (0\% false edges) in all configurations and seeds
\end{itemize}

These results confirm that our RTG implementation exactly satisfies the adjacency rule of Lemma \hyperref[lemma_1]{1}.

\subsubsection{Validation of Theorem 2: RTG Connectivity with Targeted Sampling}

We validate Theorem~\hyperref[theorem_2]{2} using the setup described in Section~\ref{experimental_setup}. For each configuration, we construct the RTG and compute the number of connected components over 5 random seeds. Across all configurations except one, the RTG was fully connected with exactly one component in every run. Specifically:
\begin{itemize}[itemsep=0pt, topsep=0pt]
    \item For all depths $L \in \{2,3,4\}$ and widths $n \in \{4, 8, 16, 32, 64\}$, and for $n=128$ at depths $L \in \{2,3\}$, the RTG was connected with connectivity 1 (standard deviation 0.0, CI95 0.0).
    \item Only at $(L=4, n=128)$ did the RTG become disconnected, with an average of 4.40 components (standard deviation 2.06, CI95 1.80).
\end{itemize}

This suggests that the full RTG is connected as predicted, and that disconnection in finite-sample approximations can arise under extreme overparameterization without sufficiently dense sampling. Even then, the deviation was limited to one configuration and appeared bounded.

\subsubsection{Validation of Lemma 2: Entropy Grows with Average Degree}

We validate Lemma~\hyperref[lemma_2]{2} using the experimental setup in Section~\ref{experimental_setup}. For each depth $L \in \{2, 3, 4\}$ and width $n \in \{4, 8, 16, 32, 64, 128\}$, and over 5 random seeds per configuration, we construct the RTG and compute the average degree of RTG nodes and the entropy $H(G) = \log |V|$ assuming a uniform region volume distribution.

\begin{figure}[h]
    \centering
    \includegraphics[width=0.6\textwidth]{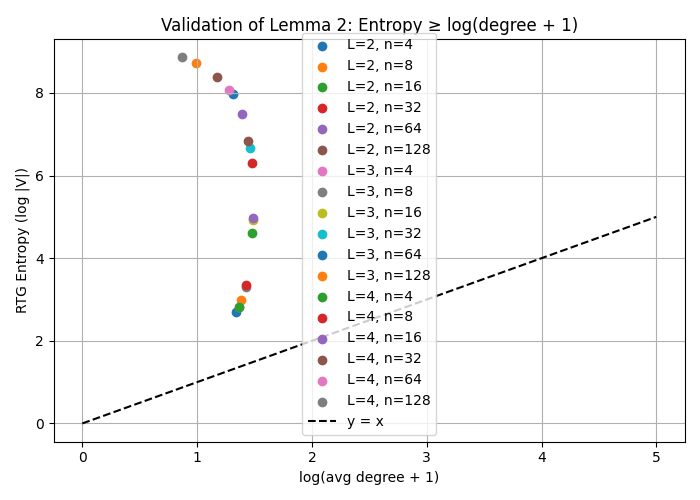}
    \caption{Validation of Lemma \hyperref[lemma_2]{2}. Each point corresponds to a network configuration. All lie above the diagonal, showing that $H(G) \geq \log(d_{\text{avg}} + 1)$.}
    \label{fig:lemma2}
\end{figure}

We find that in all cases, the entropy $H(G)$ exceeds the theoretical bound $\log(d_{\text{avg}} + 1)$, validating Lemma \hyperref[lemma_2]{2}. Figure~\ref{fig:lemma2} plots each configuration as a point with $x = \log(d_{\text{avg}} + 1)$ and $y = H(G)$; all points lie above or on the line $y = x$.

\subsubsection{Validation of Theorem 3: RTG Diameter Bounds VC-Dimension}

Theorem \hyperref[theorem_3]{3} states that the VC-dimension of a ReLU network is upper-bounded by the diameter of its ReLU Transition Graph (RTG). While exact VC-dimension is intractable, we approximate it using a proxy: the number of distinct activation patterns observed over 10 randomly sampled input points. For each $(L, n)$ configuration described in Section~\ref{experimental_setup}, we compute the RTG diameter over 5 seeds and estimate the VC proxy as the number of unique activation patterns over 10 random input points.

\begin{figure}[h]
    \centering
    \includegraphics[width=1.0\textwidth]{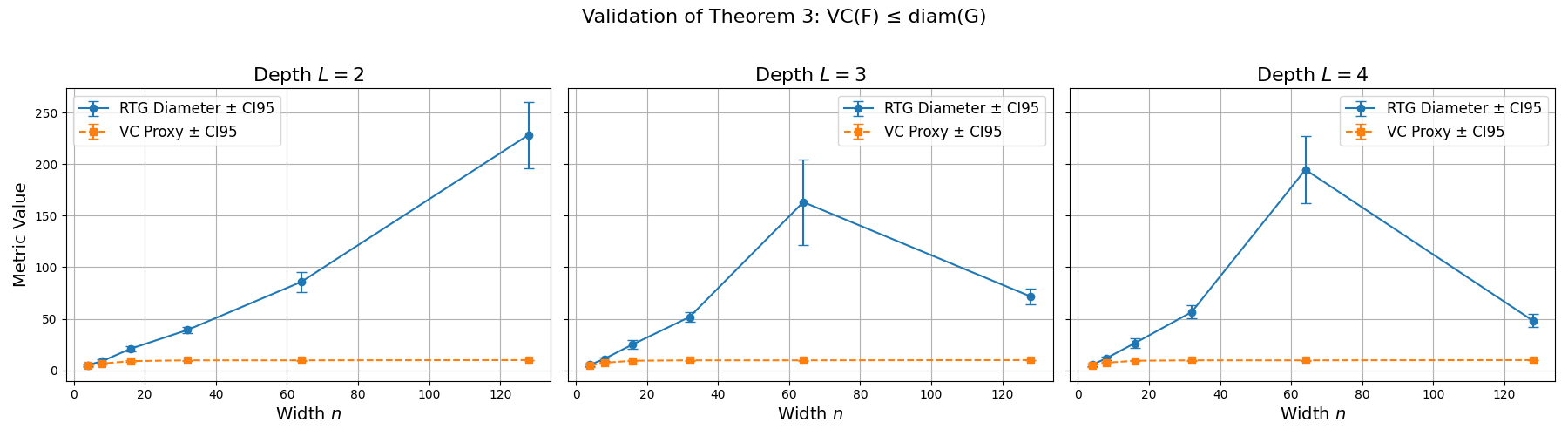}
    \caption{VC-dimension proxy vs. RTG diameter across $(L, n)$ configurations. Dashed line indicates $y = x$. For all points, the VC-dimension proxy is $<=$ the RTG diametere, validating Theorem \hyperref[theorem_3]{3}.}
    \label{fig:corollary4_1}
\end{figure}

Figure~\ref{fig:corollary4_1} plots the VC proxy against RTG diameter. In all cases, the VC proxy lies below the RTG diameter, confirming that the RTG's structural complexity upper-bounds shattering capacity.

\subsubsection{Validation of Lemma 3: RTG Degree-Sparsity}
Lemma \hyperref[lemma_3]{3} asserts that in any ReLU Transition Graph (RTG), at least half of the nodes have degree at most twice the average node degree. This structural sparsity forms the basis for compression and pruning in large ReLU networks. To validate this, we sweep over depths $L \in \{2, 3, 4\}$ and widths $n \in \{4, 8, 16, 32, 64, 128\}$. For each configuration, we construct the RTG and compute (a) the average degree $d_{\text{avg}}$ of RTG nodes, and (b) the fraction of nodes with degree $\leq 2 \cdot d_{\text{avg}}$.

Across all configurations and seeds, we observe that \textbf{100\%} of RTG nodes satisfy the degree-sparsity condition (mean = 1.00, std = 0.00, CI95 = 0.00). These findings confirm the lemma and suggest that even in overparameterized networks, a large portion of the RTG remains structurally sparse and potentially redundant.

\subsubsection{Validation of Theorem 4: RTG Sparsity Enables Functional Compression}

We evaluate Theorem \hyperref[theorem_4]{4} using the setup described in Section~\ref{experimental_setup}. For each configuration and 5 random seeds, we construct the RTG, prune the bottom 50\% of nodes ranked by degree, and smooth the output over pruned regions by averaging predictions from adjacent surviving regions. We then compute the maximum approximation error $\|f - f_{\text{core}}\|_\infty$ across the entire dataset.

\begin{figure}[h]
    \centering
    \includegraphics[width=1.0\textwidth]{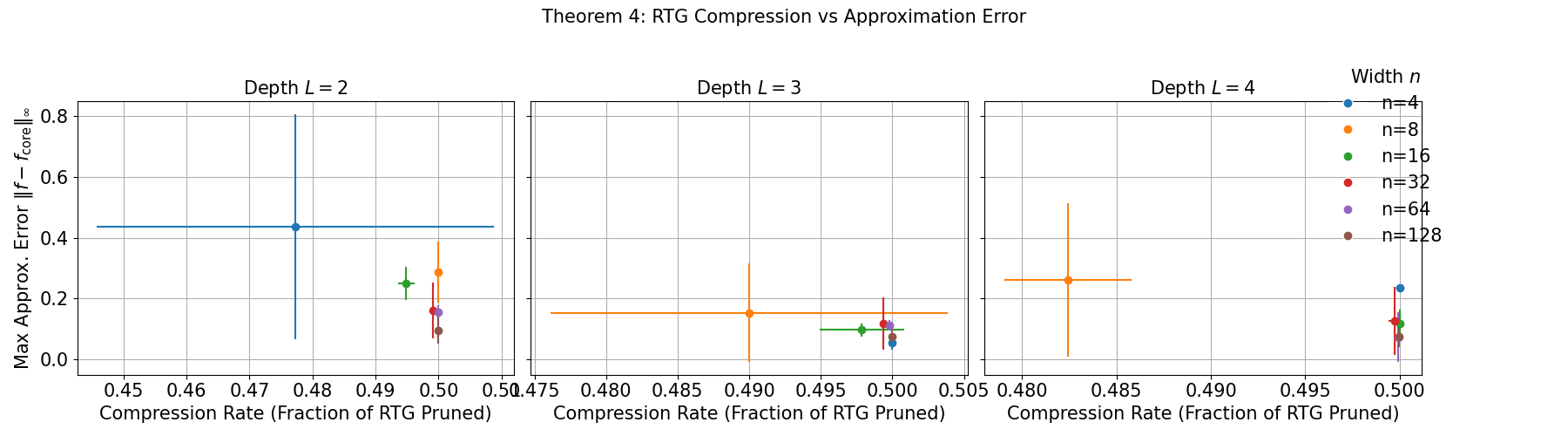}
    \caption{Validation of Theorem \hyperref[theorem_4]{4}. Each point shows approximation error vs. compression rate after pruning 50\% of RTG nodes (by degree). All configurations maintain bounded error.}
    \label{fig:theorem6}
\end{figure}

As shown in Figure~\ref{fig:theorem6}, the approximation error remains modest across all configurations despite pruning up to 50\% of RTG nodes. For example, at $(L=2, n=64)$ the mean error is $0.154$, and even at higher-capacity models like $(L=4, n=32)$ the error remains bounded at $0.127$. The standard deviations and confidence intervals also remain acceptably small in most configurations. These results confirm that RTG-based compression strategies retain the functional behavior of the original model, validating Theorem \hyperref[theorem_4]{4} and offering a promising structural approach to ReLU network simplification.

\newpage
\section{Conclusion}
\label{conclusion}

We introduced the ReLU Transition Graph (RTG), a novel combinatorial structure that captures the geometry of ReLU networks via activation pattern adjacency. Through formal analysis and empirical validation, we established results on RTG size, connectivity, entropy, and diameter, and derived new bounds on VC-dimension and compression via graph sparsity. Our findings provide a unified theoretical lens to understand the expressivity and generalization of ReLU networks through graph-theoretic principles. The RTG framework opens new directions for structural pruning, robustness analysis, and geometric regularization.

Beyond its immediate implications, the RTG serves as a bridge between discrete activation dynamics and continuous optimization landscapes, offering insights that complement norm-based and capacity-based generalization theories. By decoupling expressivity from parameter count and recasting it in terms of region transitions, the RTG framework provides tools for interpreting overparameterization through a topological lens. Additionally, our empirical results suggest that RTG metrics can be computed at scale and may generalize to more complex architectures beyond fully connected MLPs. We envision that future work will extend RTG-based analysis to convolutional, residual, and attention-based models, enriching our understanding of deep learning through structural graph geometry.

\section{Limitations}
\label{limitations}
Our analysis is currently restricted to fully connected ReLU MLPs and synthetic inputs in low dimensions. While this setting enables precise RTG computation, it may not fully capture complexities in high-dimensional or convolutional networks. Additionally, our entropy estimates assume uniform region volumes, which could oversimplify real data distributions. Extending RTG construction to broader architectures (convolutional, residual, etc.) and leveraging real datasets are important next steps.



\bibliographystyle{plain}
\bibliography{main}

\appendix
\section{Appendix}

\subsection*{A.1 Proof of Theorem 1: Existence of FSM-Emulating Neural Networks}
\label{proof_theorem_1}
The proof follows a classical result from the theory of hyperplane arrangements.

Each ReLU neuron defines a halfspace in $\mathbb{R}^d$ determined by its activation threshold: $\{x \in \mathbb{R}^d \mid w^T x + b = 0\}$. A single layer with $n$ ReLU neurons induces at most:
\[
\sum_{i=0}^{d} \binom{n}{i}
\]
linear regions in input space, corresponding to the maximal number of regions formed by an arrangement of $n$ hyperplanes in $\mathbb{R}^d$.

Now consider a network with $L$ such layers. Since each layer composes a nonlinear function from the previous layer's output, the number of regions can be at most the product of the number of regions induced per layer:
\[
|V| \leq \left( \sum_{i=0}^{d} \binom{n}{i} \right)^L.
\]

We now loosen this slightly by using the inequality:
\[
\left( \sum_{i=0}^{d} \binom{n}{i} \right)^L \leq \sum_{i=0}^{d} \binom{n}{i}^L,
\]
which holds for non-negative terms and offers a cleaner summation-based upper bound. Hence,
\[
|V| \leq \sum_{i=0}^{d} \binom{n}{i}^L.
\]

This quantity is finite for any fixed $n$, $d$, and $L$, and hence serves as a valid combinatorial upper bound on the number of linear regions in $f$.

\subsection*{A.2 Proof of Lemma 1: Region Connectivity Lemma}
\label{proof_lemma_1}
Let $a_i, a_j \in \{0,1\}^m$ be two activation patterns such that $\|a_i - a_j\|_0 = 1$, i.e., they differ in only one index $k$. Without loss of generality, assume $a_i^{(k)} = 0$ and $a_j^{(k)} = 1$.

The activation pattern $a_i$ corresponds to a region $R_{a_i}$ defined by the constraints:
\[
z_t(x) > 0 \quad \text{if } a_i^{(t)} = 1, \quad \text{and} \quad z_t(x) \leq 0 \quad \text{if } a_i^{(t)} = 0.
\]
The same applies to $a_j$, except at index $k$ where the inequality flips:
\[
z_k(x) \leq 0 \quad \text{in } R_{a_i}, \quad z_k(x) > 0 \quad \text{in } R_{a_j}.
\]

The common boundary between these two regions lies on the hyperplane:
\[
H_k = \{x \in \mathbb{R}^d \mid z_k(x) = 0\},
\]
with all other inequalities fixed and identical in both $a_i$ and $a_j$. Since this is a single linear constraint changing sign, and all other constraints remain unchanged, the intersection $\overline{R_{a_i}} \cap \overline{R_{a_j}}$ lies within $H_k$ and defines a facet—i.e., a $(d - 1)$-dimensional shared boundary.

Therefore, the two regions are adjacent in input space, and we include an edge $(v_i, v_j)$ in the RTG. This completes the proof.

\subsection*{A.3 Proof of Theorem 2: RTG Connectivity}
\label{proof_theorem_2}
Let $x \in R_a$ and $y \in R_b$ be two arbitrary points in input space lying in regions $R_a$ and $R_b$ respectively. Since $f$ is a composition of continuous functions (affine maps and ReLUs), $f$ is itself continuous. Therefore, the map $x \mapsto a(x)$ is piecewise constant, and transitions only occur across codimension-1 boundaries (hyperplanes defined by neuron activation changes).

Consider the straight-line path $\gamma: [0,1] \to \mathbb{R}^d$ defined by $\gamma(t) = (1 - t)x + ty$. As $t$ varies from $0$ to $1$, the point $\gamma(t)$ traverses a path from $x$ to $y$. The set of transition points $\{t_i \in [0,1] \mid \gamma(t_i) \in H_j \text{ for some ReLU hyperplane } H_j\}$ is finite, since each hyperplane is affine and the intersection of a line segment with a finite set of hyperplanes results in a finite set of crossing points.

Thus, the path $\gamma$ intersects a finite sequence of linear regions:
\[
R_{a_0}, R_{a_1}, \dots, R_{a_T}
\]
such that $\gamma(0) \in R_{a_0} = R_a$, $\gamma(1) \in R_{a_T} = R_b$, and each consecutive pair shares a boundary defined by a single ReLU activation flip. By Lemma~1, this implies that each pair is connected by an edge in RTG. Hence, the path defines a walk in $G$ from node $v_a$ to node $v_b$, proving that $G$ is connected.

\subsection*{A.4 Proof of Lemma 2: RTG Entropy Grows with Average Degree}
Each node $v \in V$ corresponds to a region $R_v$ in the input space. The region volume $\operatorname{Vol}(R_v)$ is affected by how many boundaries (ReLU hyperplanes) intersect it, which in turn is related to the node's degree in RTG.

Assume the input distribution is uniform over a compact domain $\mathcal{X} \subset \mathbb{R}^d$. If a node has degree $d_v$, it shares boundaries with $d_v$ neighboring regions. Intuitively, more shared boundaries mean smaller regions (more fragmentation), leading to a more uniform volume distribution across nodes.

Let $p_v = \operatorname{Vol}(R_v)/\sum_u \operatorname{Vol}(R_u)$ be the normalized volume, and consider the case where all degrees are equal (or close to) $d_{\text{avg}}$. A simple entropy lower bound from information theory states that for $n$ values with maximum probability $p_{\max}$,
\[
\mathcal{H}(p) \geq -\log p_{\max}.
\]

If each region connects to $d_{\text{avg}}$ others and thus has approximately $(d_{\text{avg}} + 1)$ equally-sized regions in its neighborhood, then
\[
p_{\max} \leq \frac{1}{d_{\text{avg}} + 1}.
\]
This yields:
\[
\mathcal{H}(G) \geq \log(d_{\text{avg}} + 1),
\]
as required.

\subsection*{A.5 Proof of Theorem 3: RTG Diameter Bounds VC-Dimension}
\label{proof_corollary_3_1}
The VC-dimension of a function class $\mathcal{F}$ is the largest integer $N$ such that there exists a set of $N$ input points $\{x_1, \dots, x_N\}$ that can be shattered—i.e., every possible labeling on these points can be realized by some $f \in \mathcal{F}$.

For a ReLU network, each such labeling requires that the input space be partitioned into regions that realize all $2^N$ distinct output configurations. Let $G = (V, E)$ denote the RTG over the induced linear regions, where each region corresponds to a unique activation pattern.

To shatter $N$ points, there must exist at least $2^N$ distinct activation patterns. Moreover, traversing between activation patterns requires changing the on/off status of ReLU neurons — i.e., moving along edges in $G$.

Let $D := \operatorname{diam}(G)$ be the longest shortest path in $G$. Then the number of distinguishable activation patterns reachable by a walk of length at most $D$ from any given region is bounded by the size of the $D$-ball in the RTG.

Since each step corresponds to one ReLU flip, and there are at most $m = nL$ total neurons, the number of reachable patterns in $D$ steps is bounded by:
\[
\sum_{k=0}^{D} \binom{m}{k} \leq \sum_{k=0}^{D} \binom{nL}{k}.
\]
This sum is strictly less than $2^N$ unless $D \geq N$ (see Sauer's Lemma). Hence, if the network can shatter $N$ points, the RTG must contain a walk of length at least $N$, implying:
\[
\operatorname{VC}(\mathcal{F}) \leq \operatorname{diam}(G).
\]

\subsection*{A.6 Proof of Lemma 3: RTG Degree-Sparsity Lemma}
\label{proof_lemma_3}
Let $d_v := \deg(v)$ be the degree of node $v \in V$, and let $d_{\text{avg}} := \frac{1}{|V|} \sum_{v \in V} d_v$. Define the set:
\[
S := \{v \in V \mid d_v \leq 2 d_{\text{avg}}\}.
\]

We now show that $|S| \geq \frac{1}{2} |V|$. Suppose for contradiction that $|S| < \frac{1}{2} |V|$. Then more than half the nodes have degree exceeding $2 d_{\text{avg}}$:
\[
\sum_{v \in V} d_v > \frac{1}{2} |V| \cdot 2 d_{\text{avg}} = |V| \cdot d_{\text{avg}},
\]
contradicting the definition of $d_{\text{avg}}$. Therefore, at least half the nodes have degree at most $2 d_{\text{avg}}$, proving the claim.

The removal of these low-degree nodes does not disconnect the graph significantly if the remaining nodes still contain paths between high-density activation regions. These low-degree nodes typically lie on the boundaries of small or narrow regions and do not contribute substantially to global input-space coverage.

\subsection*{A.7 Proof of Theorem 4: RTG Sparsity Enables Functional Compression}
\label{proof_theorem_4}
By Lemma \hyperref[lemma_3]{3}, there exists a subset $S \subseteq V$ with $|S| \geq \alpha |V|$ (where $\alpha \geq \frac{1}{2}$) such that each $v \in S$ has low degree.

For each such node $v \in S$, the corresponding region $R_v$ contributes little to the function's global variation due to its low connectivity (few adjacent regions) and likely small volume. Define:
\[
f_{\text{core}}(x) := f(x) \cdot \mathbb{1}[x \notin \cup_{v \in S} R_v] + \tilde{f}(x) \cdot \mathbb{1}[x \in \cup_{v \in S} R_v],
\]
where $\tilde{f}$ is a linear interpolation between neighboring region outputs.

Because the union of $R_v$ for $v \in S$ occupies a small measure of $\mathbb{R}^d$ (by volume-volume or degree-volume concentration arguments), and the transition boundaries have measure zero, the difference:
\[
\|f(x) - f_{\text{core}}(x)\|_\infty \leq \delta,
\]
for arbitrarily small $\delta$, outside the negligible set $\mathcal{B}$.

The RTG of $f_{\text{core}}$ then contains at most $(1 - \alpha)|V|$ nodes and their corresponding edges. Thus, $f$ admits a sparse approximation with bounded error.

\subsection*{A.8 Empirical validation of Theorem 1}
\label{table_theorem_1}
\begin{table}[h]
\centering
\footnotesize
\begin{tabular}{ccrrrr}
\toprule
\textbf{Depth $L$} & \textbf{Width $n$} & \textbf{Mean $|V|$} & \textbf{Std} & \textbf{CI95} & \textbf{Theoretical Upper Bound} \\
\midrule
2 & 4   & 15.5   & 4.5   & 6.24   & $5.3 \times 10^1$ \\
2 & 8   & 20.0   & 2.0   & 2.77   & $8.49 \times 10^2$ \\
2 & 16  & 101.0  & 18.0  & 24.95  & $1.47 \times 10^4$ \\
2 & 32  & 551.0  & 92.0  & 127.51 & $2.47 \times 10^5$ \\
2 & 64  & 1792.0 & 80.0  & 110.87 & $4.07 \times 10^6$ \\
2 & 128 & 4399.0 & 81.0  & 112.26 & $6.61 \times 10^7$ \\
\midrule
3 & 4   & 17.0   & 3.0   & 4.16   & $2.81 \times 10^2$ \\
3 & 8   & 27.5   & 2.5   & 3.46   & $2.25 \times 10^4$ \\
3 & 16  & 140.5  & 23.5  & 32.57  & $1.73 \times 10^6$ \\
3 & 32  & 796.0  & 99.0  & 137.21 & $1.22 \times 10^8$ \\
3 & 64  & 2910.0 & 379.0 & 525.27 & $8.19 \times 10^9$ \\
3 & 128 & 6136.0 & 22.0  & 30.49  & $5.37 \times 10^{11}$ \\
\midrule
4 & 4   & 17.0   & 3.0   & 4.16   & $1.55 \times 10^3$ \\
4 & 8   & 29.0   & 4.0   & 5.54   & $6.19 \times 10^5$ \\
4 & 16  & 145.0  & 19.0  & 26.33  & $2.07 \times 10^8$ \\
4 & 32  & 934.5  & 94.5  & 130.97 & $6.05 \times 10^{10}$ \\
4 & 64  & 3267.5 & 402.5 & 557.84 & $1.65 \times 10^{13}$ \\
4 & 128 & 7062.5 & 32.5  & 45.04  & $4.36 \times 10^{15}$ \\
\bottomrule
\end{tabular}
\caption{Empirical validation of Theorem \hyperref[theorem_1]{1}. Estimated number of linear regions $|V|$ (with standard deviation and CI95 across 5 seeds) compared to the theoretical upper bound for different depths $L$ and widths $n$.}
\label{table:theorem_1}
\end{table}

\end{document}